\documentclass[journal]{IEEEtran}
\usepackage{cmap}
\usepackage[T1]{fontenc}
\usepackage{graphicx}
\usepackage{epsfig}
\usepackage{amsmath}
\usepackage{amssymb}
\usepackage{amsfonts}
\usepackage{times}
\usepackage[numbers]{natbib}
\usepackage{multicol}
\usepackage[bookmarks=true]{hyperref}
\usepackage{url}
\usepackage{fullpage}
\usepackage[noend,ruled,vlined,linesnumbered]{algorithm2e}
\usepackage[font={it}]{caption}
\usepackage[caption = false]{subfig}
\usepackage{amsthm}
\usepackage{graphicx,wrapfig}
\usepackage{etoolbox}
\usepackage{array}
\usepackage{makecell}
\usepackage{mathtools}

\def\all{all}
\def\files{all}
\ifx\files\all  \else   \fi

\newcommand{\coreset}{\textsc{Coreset}}
\newcommand{\lcoreset}{\textsc{Coreset per Layer}}
\def\c{x}
\newcommand{\norm}[1]{\left\lVert#1\right\rVert}                  % Norm
\newcommand{\pr}{\mathrm{pr}}
\newcommand{\CC}{X}

\newcommand{\REAL}{\ensuremath{\mathbb{R}}}                       % REAL
\newcommand{\range}{\mathrm{range}}
\newcommand{\ranges}{\mathrm{ranges}}
\newcommand{\br}[1]{\left\{#1\right\}}                            % {#}

\newcommand{\ff}{f}

\newcommand{\ball}[1]{\mathbb{B}_{#1}(0)}

\newcommand{\ignore}[1]{}

\newtheorem{theorem}{Theorem}

\newtheorem{corollary}[theorem]{Corollary}
\newtheorem{definition}[theorem]{Definition}
\DeclareMathOperator{\RELU}{ReLU}
\DeclareMathOperator{\binary}{binary}
\DeclareMathOperator{\softclip}{soft-clipping}
\DeclareMathOperator{\gauss}{Gaussian}
\usepackage{amsmath,amsfonts,bm}

\def\eqref#1{equation~\ref{#1}}
\def\1{\bm{1}}

\DeclareMathAlphabet{\mathsfit}{\encodingdefault}{\sfdefault}{m}{sl}
\SetMathAlphabet{\mathsfit}{bold}{\encodingdefault}{\sfdefault}{bx}{n}

\newcommand{\sigmoid}{\sigma}
\newcommand{\softplus}{\zeta}

\newcommand{\real}{\mathbb{R}}

\newcommand{\eps}{\varepsilon}

\begin{document}
\title{Data-Independent Structured Pruning of Neural Networks via Coresets }
\author{Ben Mussay, Daniel Feldman, Samson Zhou, Vladimir Braverman, Margarita Osadchy
\thanks{B. Mussay, D. Feldman, and M. Osadchy are with the Department of Computer Science, Univeristy of Haifa, Haifa 31905 Israel. E-mail: \texttt{bengordoncshaifa@gmail.com, dannyf.post@gmail.com, rita@cs.haifa.ac.il}}
\thanks{S. Zhou is with the Department of Computer Science, Carnegie Mellon University, Pittsburgh, IN., USA.
E-mail: \texttt{samsonzhou@gmail.com}}
\thanks{V. Braverman is with the Department of Computer Science, Johns Hopkins University, Baltimore, MD., USA.
E-mail: \texttt{vova@cs.jhu.edu}}
}

\maketitle

\begin{abstract}
Model compression is crucial for deployment of neural networks on devices with limited computational and memory resources. Many different methods show comparable accuracy of the compressed model and similar compression rates. However, the majority of the compression methods are based on heuristics and offer no worst-case guarantees on the trade-off between the compression rate and the approximation error for an arbitrarily new sample.

We propose the first efficient structured pruning algorithm with a provable trade-off between its compression rate and the approximation error for any future test sample. Our method is based on the coreset framework and it approximates the output of a layer of neurons/filters by a coreset of neurons/filters in the previous layer and discards the rest. We apply this framework in a layer-by-layer fashion from the bottom to the top. Unlike previous works, our coreset is data independent, meaning that it provably guarantees the accuracy of the function for any input $x\in \mathbb{R}^d$, including an adversarial one.
\end{abstract}

\begin{IEEEkeywords}
Model Compression; Network Pruning; Structured Pruning; Coreset
\end{IEEEkeywords}%}

%\IEEEdisplaynontitleabstractindextext

% \fi
%\IEEEpeerreviewmaketitle

%\IEEEpeerreviewmaketitle

\section{Introduction}
Neural networks today are the most popular and effective instrument of machine learning with numerous applications in different domains. Since~\cite{krizhevsky2012imagenet} used a model with 60M parameters to win the ImageNet competition in 2012, network architectures have been growing wider and deeper. The vast overparametrization of neural networks offers better convergence~\citep{Allen_ZhuLS19} and better generalization~\citep{LeCunSrebro}. The downside of the overparametrization is its high memory and computational costs, which prevent the use of these networks in small devices, e.g., smartphones. Fortunately, it was observed that a trained network could be reduced to smaller sizes without much accuracy loss. Following this observation, many approaches to compress existing models have been proposed (see~\cite{sparcification_review} for a recent review  on network sparsification, and~\cite{mozer1989skeletonization, SrivastavaHKSS14,NISP,HeZS17, dcp, median-pruning, GateDG} for structured pruning).

Network compression is composed of two parts: 1) finding a more compact architecture (reducing the size of the layers or reducing the number of non-zero weights) and 2) updating the network parameters to approximate the original model.
Most of the model compression methods search for a smaller architecture and weight approximation simultaneously~\cite{median-pruning, dcp, slim}. A different approach is to decouple architecture search from the network approximation (e.g.~\citep{nec,rethinking}). A lot of work has been done in neural architecture search~\cite{ElskenMH19}. One can follow this route for finding a smaller architecture~\citep{automl}.
%In this paper, we focus on finding the best approximation of the original network, given the target smaller architecture.
The second step of determining the weights of the small model can be done in several ways:
\begin{itemize}
    \item Train the small architecture from scratch -- the experiments in \citep{rethinking} showed that this could reach accuracy similar to the original large network.
    \item  Retain $K$ most important weights/filters ($K$ is defined by the small architecture)  and fine-tune the network (e.g.~\cite{nec}).
    \item  Find the weights of the small architecture that approximate the output of the original network and then fine-tune the weights (e.g.,~\cite{HeZS17,thin}).
\end{itemize}

Even though many approaches show comparable results in terms of accuracy of the compressed model and its size, there are other important aspects that a practitioner should consider. For example, if the algorithm is tuned to a specific data (as in~\citep{mozer1989skeletonization, SrivastavaHKSS14,network_trimming,NISP,baykal,dcp}), the compressed network would fail to approximate the output of the original network on less typical unseen examples. In contrast, the accuracy of data-independent method (e.g.~\cite{nec,median-pruning}) is guaranteed for any input. Another important aspect of a compression framework is its construction time. Specifically, one-shot compression~\cite{nec} is much faster than an iterative approach (e.g.~\cite{dcp}).  Additionally, a well approximated model would require shorter fine-tuning time than a less accurate counterpart or training the compressed architecture from scratch. Obviously, the approximation of the original model is a function of the compression rate. However, previous works generally lack strong provable guarantees on the \textbf{trade-off between the compression rate and the approximation error}. The absence of worst-case performance analysis can potentially be a glaring problem depending on the application. Finally, there is a choice between model sparcification~\cite{lebedev2016fast, dong2017learning} vs. structured pruning~\cite{network_trimming,Channel,filter_pruning,slim}. The former, leads to an irregular network structure, which needs a special treatment to deal with sparse representations, making it hard to achieve actual computational savings.  Structured pruning simply reduces the size of the tensors, providing very simple implementation and much higher computational savings. Thus structured pruning is much more practical.

Form the above discussion, we can derive the following requirements for the model compression framework:  1) it should provide provable guarantees on the trade-off between the compression rate and the approximation error, 2) it should be data-independent, 3) it shout yield high compression rates, 4) it should be computationally efficient.
%%%%%%%%%%%upto here.

To address these goals, we propose an efficient framework with provable guarantees for structured pruning, which is based on the existing theory of coresets~\citep{BravermanFL16}. Coresets decrease massive inputs to smaller instances while maintaining a good provable approximation of the original set with respect to a given function. While most of previous work used coreset to reduce large volumes of data, here we apply it to reduce model parameters in neural networks by treating neurons as inputs in a coreset framework. Namely, we reduce the number of neurons in layer $i$ by constructing a coreset of neurons in this layer that provably approximates the output of neurons in layer $i+1$ and discarding the rest. The coreset algorithm provides us with the choice of neurons in layer $i$ and with the new weights connecting these neurons to layer $i+1$.  We apply similar concept to derive a filter pruning algorithm for the convolutional layers. The coreset algorithm is applied layer-wise from the bottom to the top of the network, resulting in a unified framework for approximating the entire network.

Our framework satisfies the aforementioned requirements:

\noindent \textbf{Provable guarantees:} The size of the coreset, and consequently the number of remaining neurons in layer $i$, is provably related to the approximation error of the output for every neuron in layer $i+1$. Thus, we can theoretically derive the trade-off between the compression rate and the approximation error of any layer in the neural network. We implement the direction, in which we construct a coreset of a predefined size (the size is specified by the target small architecture).\footnote{The other direction, in which the size of the coreset is determined given the approximation error is theoretically possible, but is more complicated to implement.} and provably derive the approximation error, which holds for any input. Most previous work lack such guarantees, and thus shows abnormal behaviour on some of the inputs. It was shown in~\cite{brain_workshop} that pruning via recent methods affects generalization of a network at a class and exemplar level. Full and pruned models have comparable average accuracy, but they diverge considerably
in behavior on specific sub-sets of input distribution. Our framework does not suffer such a degradation as it provably holds for \emph{any} input.
%Using coreset theoretical framework we guarantee provable layer-vise approximation of the network. The approximation error of each neuron/map is approximately Gaussian, thus  most of it is filtered out by the following layers(as was shown in~\citep{Arora}) resulting in accurate approximation of the entire network.

\noindent \textbf{Data-independent compression:}
The coreset approximation of neurons provably holds for any input, in particular, it holds for the input layer; thus the proposed framework will provably hold for any input data, termed data-independent compression.

\noindent \textbf{Fast construction:}
Our framework performs one-shot structured pruning that seeks for the best approximation of the original network given the architecture of the small network.  Thus it requires much shorter fine-tuning than heuristic-based approximation with no guarantees or training from scratch.

\noindent \textbf{High Compression Rates:}
The proposed theory shows the existence of a small neural coreset. Our empirical results on LeNet-300-100 for MNIST~\citep{MNIST} and VGG-16~\citep{Simonyan14c} for CIFAR-10~\citep{krizhevsky2009learning} demonstrate that our framework based on coresets of neurons outperforms sampling-based coresets by improving compression without sacrificing the accuracy. The results on VGG-19, ResNet56 \citep{resnet} for CIFAR-10 and  ResNet50 for ImageNet \citep{ILV} demonstrate that our channel pruning method shows comparable results to the state-of-the-art methods of channel pruning, but is more efficient in terms of model construction.

\section{Related Work}
State-of-the-art neural networks are often overparameterized, which causes a significant redundancy of weights.  To reduce both computation time and memory requirements of trained networks, many approaches aim at removing this redundancy by model compression. These works can be categorized based on different criteria, for instance, sparcification (reducing the number of non-zero weights) vs. structured pruning (reducing the width of the layer); data-dependent (trainable) vs. data independent (that works for any input); one-shot pruning vs. iterative pruning. Below, we discuss previous work following some of these categories.

\subsection{Sparsification vs. Structured Methods}
\textbf{Sparsification}, also known as weight pruning, was considered as far back as 1990~\citep{lecun1990optimal}, but has recently seen more study~\citep{lebedev2016fast, dong2017learning, TheLT, rewind}. One of the most popular approaches is pruning via sparsity. Sparsity can be enforced by $L_1$ regularization to push weights towards zero during training~\citep{network_trimming}. However, it was observed~\citep{Han} that after fine-tuning of the pruned network, $L_2$ regularized network outperformed $L_1$, as there is no benefit to pushing values towards zero compared to pruning unimportant (small weight) connections.

The approach in~\cite{denton} exploits the linearity of the neural network by finding a low-rank approximation of the weights and keeping the accuracy within 1\% of the uncompressed model. Quantization of the neural network's weights are done recently for saving memory. One example of such an approach,  proposed in \citep{Google}, performs quantization of the neural network's weights and suggests a new training procedure to preserve the model accuracy after the quantization.

Sparification methods showed high reduction in non-zero weights, e.g., the compression rate of AlexNet can reach 35x with the combination of pruning, quantization, and Huffman coding \citep{Han-ICLR}.

The main drawback in weight pruning is that it leads to an irregular network structure, which needs a special treatment to deal with sparse representations, making it hard to achieve actual computational savings.

\textbf{Structured Pruning} (e.g.,~\citep{network_trimming,Channel,filter_pruning,slim,dcp,nec}) simply reduces the size of the tensors.  In the case of CNNs it reduces the number of filters in a layer. These methods are preferable over sparsification, as the resulting models are very easy to use and they save an actual inference time.

The method in~\citep{filter_pruning} measures the importance of channels by calculating the sum of absolute values of weights. Other channel pruning methods either impose channel-wise sparsity in training, followed by pruning channels with small scaling factors, and fine-tuning (e.g,~\citep{slim})
or perform channel pruning by minimizing the reconstruction error of feature maps between the pruned and pre-trained model (e.g.,~\citep{HeZS17}.) The method in \citep{GateDG} computes the impact of filter removal using Taylor expansion.
%These methods lack provable guarantees on the trade-offs between their accuracy and compression.

Our method belongs to structured pruning, and thus the resulting pruned networks provide large savings in computations and are easy to implement.

\subsection{Data-Dependent vs. Data-Independent}
\textbf{Data-dependent} pruning approaches~\citep{slim,thin,PruningCN,baykal,dcp}  utilize training/validation data to determine the filters to be pruned.
The method in~\citep{network_trimming} first identifies weak neurons by analyzing their activation on a large validation dataset. Then those weak neurons are pruned and the network is retrained. The processes are repeated several times. The method in~\cite{Channel} evaluates the contribution of filters to the discriminative power of the network by utilizing training/validation data. The method in \citep{collab} captures the dependency between channels in each layer using Taylor expansion and then exploits these dependencies to determine which channels to prune. The optimization of the channel selection problem requires evaluating Hessian on training samples.

\textbf{Data-independent} methods~\cite{nec,soft,YeL0W18,median-pruning} compress neural network based on weights' characteristics. The work in~\cite{median-pruning} observes that keeping only the filters with the largest norm (as was done in ~\cite{nec,soft,YeL0W18}) does not provide a good approximation of the network and the distribution of norms in each layer. To resolve this problem, \cite{median-pruning} suggests to  search for filters with the most replaceable contribution. These are found by the means of Geometric Median.

Data-independent pruning increases the robustness of the compressed network to future samples, unlike data-dependent pruning, which utilizes the training data. Our method is data-independent and it is theoretically guaranteed to maintain the same accuracy on the out-of-the-distribution samples.

\subsection{Three Criteria: Accuracy, Compression, Efficient Construction }
It was pointed out in~\citep{renda}, that network compression should be evaluated based on three criteria: accuracy, inference efficiency, and construction efficiency. Accuracy has been the main target in all previous work. As we discussed in the previous section, inference efficiency is not equivalent to the compression rate: Structured methods that provide the same compression rate as the sparsification methods yield much higher computational savings in inference time.

The last criterion -- construction efficiency, includes both the complexity of the pruning algorithm (for example one-shot methods are much less expensive than the iterative methods) and the amount of training or fine-tuning required. Most previous work have neglected this criterion. Furthermore, \citep{rethinking} observed that fine-tuning a pruned model gives a comparable performance with training that model with randomly initialized weights, concluding that pruning algorithms that assume a predefined target network architecture could be replaces with a direct training of the target architecture from scratch. We show that this suggestion has a profound drawback, which is the efficiency of producing the small network. Fine-tuning the approximated model is much faster than training the same architectures from random initialization. We show this empirically on several architectures in Section~\ref{sec_experiments_CP}. Thus both architectural search and network approximation play an important role in network compression.

% Using coreset theoretical framework we guarantee provable layer-vise approximation of the network. Since the approximation error of a layer is approximately Gaussian, most of this noise is filtered out by the following layers (as was shown in~\citep{Arora}). Since our method  approximates the large model with a smaller one, the approximated model is much closer to the original solution than a random network of the same size.

%Our method is one-shot and the construction algorithm is very efficient. Together with rapid convergence in fine-tuning, our method excels in the third criterion as well.

\subsection{Coresets}
Our compression algorithm is based on a data summarization approach known as coresets.
Over the past decade, coreset constructions have been recognized for high achievements in data reduction in a variety of applications, including $k$-means, SVD, regression, low-rank approximation, PageRank, convex hull, and SVM; see details in~\cite{phil}. Many of the non-deterministic coreset based methods rely on the sensitivity framework, in which elements of the input are sampled according to their sensitivity~\citep{LangbergS10, BravermanFL16, Elad}, which is used as a measure of their importance. The sampled elements are usually reweighted afterwards.

\subsection{Coreset-based Model Compression}
Similar to our work, the approachs in~\citep{baykal,Liebenwein2020Provable} use corests for model compression. However, \citep{baykal} constructs  coresets of weights, while we construct  coresets of neurons. Both~\citep{baykal,Liebenwein2020Provable}  compute the importance of  weight/filter, using a validation set. The coreset is chosen for the specific distribution (of data) so consequently, the compressed model is data-dependent. In our construction, the input of the neural network is assumed to be an arbitrary vector in $\mathbb{R}^d$ and the sensitivity of a neuron is computed for every input in $\mathbb{R}^d$.  This means that we create a data-independent coreset; its size is independent of the properties of the specific data at hand, and the compression provably approximates any future test sample.

In \citep{DubeyCA18}, k-means coresets were suggested to compress layers by adding a sparsity constraint. The weighting of the filters in the coreset was obtained based on their activation magnitudes over the training set. The compression pipeline also included a pre-processing step that followed a simple heuristic that eliminates filters based on the mean of their activation norms over the training set. This construction is obviously data-dependent and it uses corsets as an alternative mechanism for low-rank approximation of filters.

\section{Method}

We propose an algorithm for compressing layer $i$ and we apply it to all layers from the bottom to the top of the network. We start with a necessary background on coresets (Section \ref{sec:backgorund}). We continue with an intuitive description of the algorithm for the fully connected layers (Section~\ref{sec:intuation}).  We then formalize it and provide a theoretical analysis of the proposed construction (Section~\ref{sec:analysis}). Finally, we extend our theoretical results to a convolution layer (Section~\ref{sec:conv_layer}).

\begin{figure}[h!]
\centering
\subfloat[]{
\includegraphics[width=0.52\textwidth]{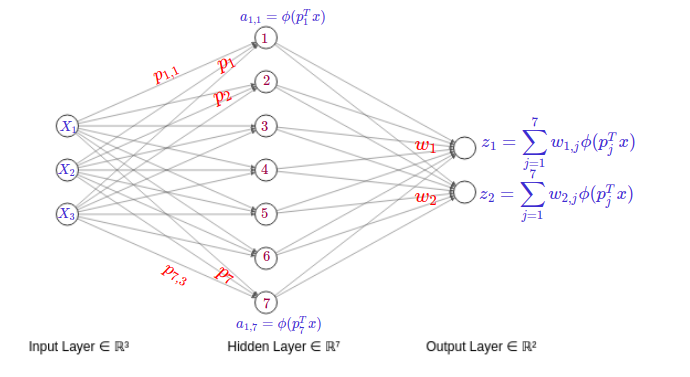}}\\
\subfloat[]{
\includegraphics[width=0.52\textwidth]{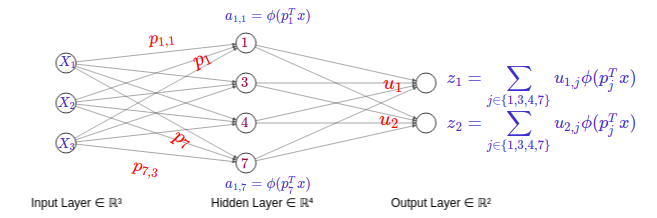}}
\caption{Illustration of our neuron coreset construction on a toy example: (a) a full network, (b) the compressed network. Both neurons in the second layer in (b) choose the same coreset comprising neurons $\{1,2,3,7\}$  from layer 1, but with different weights. The compressed network has pruned neurons $\{2,5,6\}$ from layer 1.} \label{fig:schem_single_n}
\end{figure}

\subsection{Background}\label{sec:backgorund}
\begin{definition}[weighted set]
Let $P\subset \REAL^d$ be a finite set, and $w$ be a function that maps every $p\in P$ to a \emph{weight} $w(p)>0$.
%That is, $w:X'\to[0,\infty)$ for some $X'\supseteq P$.
The pair $(P,w)$ is called a \emph{weighted set}.
\end{definition}
%A coreset is a small weighted subset of an input set, which is usually a larger weighted set.
A coreset in this paper is applied to a query space, which consists of an input weighted set, an objective function, and a class of models (queries) as follows.
\begin{definition}[Query space]\label{def::query space}
Let $P'=(P,w)$ be a weighted set called the \emph{input set}. Let $\CC\subseteq\REAL^d$ be a set of queries, and $\ff:P\times \CC\to [0,\infty)$ be a \emph{loss function}. The tuple $(P,w,\CC,\ff)$ is called a \emph{query space}.
%For every $x\in \CC$ we define the overall fitting error of $P'$ to $x$ by
%$f(P',x)=\sum_{p\in P}w(p)f(p,x)$.
\end{definition}

Given a set of points $P$ and a set of queries $\CC$, a coreset of $P$ is a weighted set of points that provides a good approximation to $P$ for any query $x\in\CC$.
We state the definition of coresets with multiplicative guarantees below, though we shall also reference coresets with additive guarantees.
\begin{definition}[$\varepsilon$-coreset, multiplicative guarantee]
Let $(P, w,\CC, \ff)$ be a query space, and $\varepsilon \in (0, 1)$ be an error parameter.
An $\varepsilon$-coreset of $(P, w, \CC, \ff)$ is a weighted set $(Q, u)$ such that for every $x \in \CC$
\[\left|\sum_{p \in P}w(p)\ff(p, x) - \sum_{q \in Q}u(q)\ff(q, x)\right| \leq \varepsilon \sum_{p \in P}w(p) \ff(p, x)\]
\end{definition}

The size of our coresets depends on two parameters: the complexity of the activation function, which is defined below, and the sum of a supremum that is defined later.
We now recall the well-known definition of VC dimension~\citep{vapnik2015uniform} using the variant from~\citep{Sensitivity}.
\begin{definition}[VC-dimension~\citep{Sensitivity}\label{vdim}]
Let $(P,w,X,f)$ be a query space.
    For every $x\in \REAL^d$, and $r\geq 0$ we define
    $$\range_{P,\ff}(\c,r):=\br{p\in P\mid \ff(p,\c)\leq r}$$
    and
      \begin{align}
      \nonumber&\ranges(P,\CC,f):=\\
      \nonumber&\br{C\cap \range_{P,\ff}(\c,r)\mid C\subseteq P, \c\in \CC, r\geq 0}.
      \end{align}
 For a set $\ranges$ of subsets of $\REAL^d$, the VC-dimension of $(\REAL^d,\ranges)$ is the size $|C|$ of the largest subset $C\subseteq \REAL^d$ such that
    \[
    |\br{C\cap \range \mid \range\in\ranges}|= 2^{|C|}.
    \]
    The \emph{VC-dimension of the query space} $(P,\CC,\ff)$ is the VC-dimension of $(P,\ranges(P,\CC,f))$.
\end{definition}

The VC-dimension of all the query spaces that correspond to the activation functions in Table~\ref{compression_rates} is $O(d)$, as most of the other common activation functions~\citep{anthony2009neural}.
\ignore{
\begin{definition}[simple operations\label{op}]
    Let $f$ be a function that can be evaluated by an algorithm that uses no more than $z$ of the following operations:
      \begin{enumerate}%[topsep=0pt,itemsep=0ex,label=(\roman*)]
      \item the exponential function $\alpha\mapsto e^\alpha$ on real numbers,
      \item the arithmetic operations $+, -, \times, $ and $/$ on real numbers,
      \item jumps conditioned on $>,\geq, <, \leq, =,$ and $\neq$ comparisons of real numbers.
      \end{enumerate}
      then we say that the function $f$ \emph{can be evaluated using $O(z)$ simple operations}.
    \end{definition}

    The following corollary is a simple variant of~\cite[Theorem 8.14]{anthony2009neural}, which was proved in~\cite[Corollary 7.4]{zahi}.
        \begin{corollary}\label{pdi}%~\cite[Theorem 8.14]{anthony2009neural}]
    Let $X\subseteq\REAL^d$ and $f:P\times X\to[0,\infty)$.
    Suppose that the value $f(p,x)$ can be computed in $O(z)$ simple operations, for every $p\in P$ and $x\in X$.
Then the VC-dimension of $(P,X,f)$ is $O(d(d+z))$.
    \end{corollary}
}

The following theorem bounds the size of the coreset for a given query space and explains how to construct it.
Unlike previous papers such as~\citep{Sensitivity}, we consider additive error and not multiplicative error.
\begin{theorem}[Coreset Construction \cite{BravermanFL16}\label{cor11}]
Let $d$ be the VC-dimension of a query space $(P,w,X,f)$.
Suppose $s:P\to[0,\infty)$ such that $s(p)\geq w(p)\sup_{x\in X}f(p,x)$.
Let $t= \sum_{p\in P}s(p)$, and $\eps,\delta\in(0,1)$. % be the total importance for an $m$-reweight of $(P,w,Q,f)$.
Let $c\geq1$ be a sufficiently large constant that can be determined from the proof, and let $C$ be a sample (multi-set) of
\[
m\geq \frac{ct}{\eps^2}\left(d\log t+\log\left(\frac{1}{\delta}\right)\right)
\]
i.i.d. points from $P$, where for every $p\in P$ and $q\in C$ we have $\pr(p=q)= s(p)/t$. Then, with probability at least $1-\delta$,
\[
\forall x\in X: \left|\sum_{p\in P} w(p)f(p,x)-\sum_{q\in C} \frac{w(q)}{m\pr(q)}\cdot f(q,x)\right| \leq \eps.
\]
\end{theorem}

\begin{table}[!h]
\centering
%\rowcolors{2}{gray!10}{}
\begin{tabular}{||c c c||}
 \hline
 Activation Function & Definition &\\
 \hline\hline
 $\RELU$ & $\max(x, 0)$ & \\[1.1ex]
 $\sigmoid$ & $\frac{1}{1 + e^{-x}}$ & \\[1.1ex]
 $\binary$ & $\begin{cases}
	0 & \text{for } x < 0\\
	1 & \text{for } x \ge 0\end{cases}$ &\\[1.1ex]
 $\softplus$ & $\ln(1 + e^x)$ & \\[1.1ex]
 $\softclip$ & $\frac{1}{\alpha}\log\frac{1+e^{\alpha x}}{1+e^{\alpha(x-1)}}$ & \\[1.1ex]
 $\gauss$ & $e^{-x}$ & \\[1.1ex]
 \hline
\end{tabular}
\vspace{0.05cm}
\caption{Examples of activation functions $\phi$ for which we can construct a coreset of size  $O(\frac{\alpha\beta}{\eps ^ 2})$ that approximates $\frac{1}{|P|}\sum_{p\in P}\phi(p^Tx)$  with $\eps$-additive error.}
\label{compression_rates}
\end{table}

\subsection{Data-Independent Coreset for Neural Pruning}\label{sec:intuation}
Let $a_j^i=\phi(p_j^Tx)$ be the $j$th neuron in layer $i$, in which $p_j$ denotes its weights and $x$ denotes an arbitrary input in $\mathbb{R}^d$ (see Figure~\ref{fig:schem_single_n}, top). We first consider a single neuron in layer $i+1$. The linear part of this neuron is $z= \sum_{j=1}^{|P|} w(p_j)\phi(p_j^T x)$. We would like to approximate $z$ by  $\tilde{z}=\sum_{l\in J^*} u(p_l)\phi(p_l^T x)$ where  $J^*\subset\{1,...,|P|\}$ is a small subset, and we want this approximation to be bounded by a  multiplicative factor that holds for any $x\in\mathbb{R}^d$.
%with  a bounded multiplicative error that holds for any $x\in\mathbb{R}^d$, such that  $J^*\subset\{1,...,|P|\}$  of size $|C|$ and $C$ is a small subset of $P$.
Unfortunately, our result in Theorem~\ref{lem:elad} shows that this idealized goal is impossible. However, we show in  Theorem~\ref{lem:sensitivity} and Corollary~\ref{corollary::negative-coreset} that we can construct a small coreset $C$, such that  $|z-\tilde{z}|\leq \eps$ for any input $x\in\mathbb{R}^d$.

Algorithm~\ref{coreset} summarizes the coreset construction for a single neuron with an activation function $\phi$,  (our results for common neural activation functions  are summarized in Table~\ref{compression_rates}).
Algorithm~2 and Corollary~\ref{corollary::double-coreset} show the construction of a single coreset with possibly different weights for all neurons in layer $i+1$ (see Figure~\ref{fig:schem_single_n}, bottom).  %The Coreset algorithms  outputs a subset of neurons and a new weight matrix for layer $i+1$.
%The approximation error of the layer becomes $O(\eps)$. Finally, for the fully connected network with $L$ layers, the approximation error of the compressed network is $O(L\eps)$ which holds for any input in $\mathbb{R}^n$.

\begin{algorithm}[!htb]
\caption{$\coreset(P,w,m,\phi,\beta)$  }\label{coreset}
{\begin{tabbing}
\textbf{Input:} \quad\quad \=A weighted set $(P,w)$, \\\>an integer (sample size) $m \geq 1$,
\\ \>an (activation) function $\phi:\REAL\to[0,\infty)$,
\\ \> an upper bound $\beta>0$.\\
\textbf{Output:} \>A weighted set $(C,u)$;
\\ \>see Theorem~\ref{lem:sensitivity} and Corollary~\ref{corollary::negative-coreset}.
\end{tabbing}}
%\vspace{-0.3cm}
%\For{every $p\in P$}
%{
%$\displaystyle s(p):= $
%}
\For{every $p\in P$}
{
$\pr(p):=\displaystyle\frac{w(p)\phi(\beta \norm{p})}{\sum_{q\in P}w(q)\phi(\beta \norm{q})}$\\
$u(p):=0$
}
$C \gets \emptyset$
\For{$m$ iterations}
{
Sample a point $q$ from $P$ such that $p\in P$ is chosen with probability $\pr(p)$.\\
$C:=C\cup\{q\}$\\
$u(q):= u(q)+\displaystyle\frac{w(q)}{m\cdot\pr(q)}$
}
\Return $(C,u)$
\end{algorithm}

\subsection{Main Theoretical Results}\label{sec:analysis}

\begin{figure}%[hb!]
\centering
\includegraphics[width=0.5\textwidth]{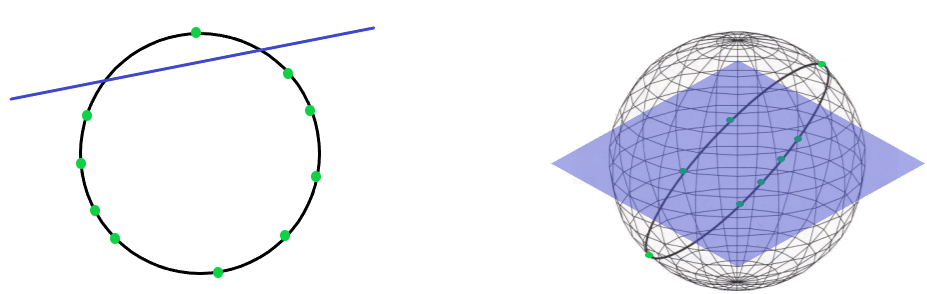}
\caption{\textbf{(left) } Any point on a circle can be separated from the other points via a line.
\textbf{(right) }The same holds for a circle which is the intersection of a $d$-dimensional sphere and a hyperplane; see Theorem~\ref{lem:elad}.}\label{fig2}
\end{figure}

Most of the coresets provide a $(1+\eps)$-multiplicative factor approximation for every query that is applied on the input set. The bound on the coreset size is independent or at least sub-linear in the original number $n$ of points, for any given input set.
Unfortunately, the following theorem proves that it is impossible to compute small coresets for many common activation functions including ReLU. This holds even if there are constraints on both the length of the input set and the test set of samples.
\begin{theorem}[No coreset for multiplicative error]
\label{lem:elad}
Let $\phi:\REAL\to[0,\infty)$ such that $\phi(b)>0$ if and only if $b> 0$.
Let $\alpha,\beta>0$, $\eps\in (0,1)$ and $n\geq1$ be an integer. Then there is a set $P\subseteq \ball{\alpha}$ of $n$ points such that if a weighted set $(C,u)$ satisfies $C\subseteq P$ and
%\begin{equation}\label{lower}
\begin{align}\label{lower}
&\forall x\in \ball{\beta}: \\
\nonumber &\left|\sum_{p\in P}\phi(p^Tx)-\sum_{q\in C}u(q)\phi(q^Tx)\right|\leq \eps\sum_{p\in P}\phi(p^Tx),
%\end{equation}
\end{align}
then $C= P$.
\end{theorem}
\begin{proof}
Consider the points on $\ball{\alpha}$ whose norm is $\alpha$ and last coordinate is $\alpha/2$.
This is a $(d-1)$-dimensional sphere $S$ that is centered at $(0,\cdots,0,\alpha/2)$.
For every point $p$ on this sphere there is a hyperplane that passes through the origin and separates $p$ from the rest of the points in $S$. Formally, by applying the Hyperplane separation theorem, there is an arbitrarily short vector $x_p$ (which is orthogonal to this hyperplane) such that $x_p^Tp>0$, but $x_p^Tq<0$ for every $q\in S\setminus \br{p}$; see Fig.~\ref{fig2}. By the definition of $\phi$, we also have $\phi(x_p^Tp)>0$, but $\phi(x_p^Tq)=0$ for every $q\in S\setminus \br{p}$.

Let $P$ be an arbitrary set of $n$ points in $S$, and $C\subset P$. Hence exists a point $p\in P\setminus C$. By the previous paragraph,
\begin{align*}
\nonumber&\left|\sum_{q\in P}\phi(x_p^Tq)-\sum_{q\in C}u(q)\phi(x_p^Tq)\right|\\&=\left|\phi(x_p^Tp)-0\right|=\phi(x_p^Tp)\\
\nonumber &=\sum_{q\in P}\phi(x_p^Tq)>\eps \sum_{q\in P}\phi(x_p^Tq).
\end{align*}
Therefore $C$ does not satisfy~\eqref{lower} in Theorem~\ref{lem:elad}.
\end{proof}

The following theorem motivates the usage of additive $\varepsilon$-error instead of multiplicative $(1+\varepsilon)$ error. Fortunately, in this case there is a bound on the coreset's size for appropriate sampling distributions.%Each input point in the following theorem represents a neuron
\begin{theorem}
\label{lem:sensitivity}
Let $\alpha,\beta>0$ and $(P,w,\ball{\beta},f)$ be a query space of VC-dimension $d$ such that $P\subseteq \ball\alpha$, the weights $w$ are non-negative, $f(p,x)=\phi(p^Tx)$ and $\phi:\REAL\to[0,\infty)$ is a non-decreasing function.
Let $\eps,\delta\in(0,1)$ and
\[
m\geq \frac{ct}{\eps^2}\left(d\log t+\log\left(\frac{1}{\delta}\right)\right)
\]
where $$t = \phi(\alpha \beta)\sum_{p\in P}w(p)$$ and $c$ is a sufficiently large constant that can be determined from the proof.

Let $(C,u)$ be the output of a call to $\coreset(P,w,m,\phi,\beta)$; see Algorithm~\ref{coreset}. Then, $|C|\leq m$ and, with probability at least $1-\delta$,
\[
\left|\sum_{p\in P}w(p)\phi(p^Tx)- \sum_{p\in C}u(p)\phi(p^Tx)\right| \leq \eps.
\]
\end{theorem}

\begin{proof}
We want to apply Algorithm~\ref{coreset}, and to this end we need to prove a bound that is independent of $x$ on the supremum $s$, the total supremum $t$, and the VC-dimension of the query space.

\noindent\textbf{Bound on $f(p,x)$:} Put $p\in P$ and $x\in \ball\beta$. Hence,
\begin{align}
    \label{ff1}f(p,x)=\phi(p^T x)&\leq \phi(\norm{p} \norm{x})  \\
    \label{ff3}&\leq  \phi(\norm{p}\beta)\\
    \label{ff2}&\leq \phi(\alpha\beta) ,
\end{align}
where~\eqref{ff1} holds by the Cauchy-Schwarz inequality and since $\phi$ is non-decreasing,~\eqref{ff3} holds since $x\in\ball{\beta}$, and~\eqref{ff2} holds since $p\in\ball{\alpha}$.
%The value of $f(p,x)$ can then be bounded by a value that is independent of $x$,
%\begin{align}
%\nonumber f(p,x)&=\frac{\phi(p^Tx)}{\lambda\sum_{q\in P}w(q)}
%%\nonumber\leq \frac{\phi(p^Tx)}{\lambda \sum_{q\in P}w(q)}\\
% \label{gg1}&\leq \frac{\phi(\norm{p}\beta)}{\lambda\sum_{q\in P}w(q)},
%\end{align}
%where the inequality is by~\eqref{ff3}. Since the last inequality holds for every $x\in \ball{\beta}$, we obtain
%\[
%s(p)=w(p)\cdot \frac{\phi(\norm{p}\beta)}{\lambda\sum_{q\in P}w(q)}\geq w(p)\sup_{x\in\ball{\beta}} f(p,x).
%\]

\noindent\textbf{Bound on the total sup $t$:}
Using our bound on $f(p,x)$,
\begin{align}
\nonumber t=\sum_{p\in P}s(p)=\sum_{p\in P}w(p)\phi(\norm{p}\beta)\leq \phi(\alpha\beta)\sum_{p\in P}w(p),
%=\frac{\phi(\alpha\beta)}{\lambda},
\end{align}
where the last inequality is by~\eqref{ff2}.

%$\displaystyle s(p):=\frac{w(p)\phi(\beta  \norm{p}) + \lambda w(p)}
%{ w(p)\phi(\beta \norm{p}) + \lambda\sum_{q \in P}{ w(q)}}$

\noindent\textbf{Bound on the VC-dimension:} of the query space $(P,w,\ball{\beta},f)$ is $O(d)$ as proved e.g. in~\citep{anthony2009neural}.

\noindent\textbf{Putting all together:} By applying Theorem~\ref{coreset} with $X=\ball{\beta}$, we obtain that, with probability at least $1-\delta$,
\[
\forall x\in \ball{\beta}: \left|\sum_{p\in P} w(p)f(p,x)-\sum_{q\in C}u(q) f(q,x)\right| \leq \eps.
\]
Assume that the last equality indeed holds. Hence,
\[
\forall x\in \ball{\beta}:\left|\sum_{p\in P} w(p)\phi(p^Tx)-\sum_{q\in C} u(q) \phi(q^Tx)\right| \leq \eps.
\]
\end{proof}

As weights of a neural network can take positive and negative values, and the activation functions $\phi:\REAL\to \REAL$ may return negative values, we generalize our result to include negative weights and any monotonic (non-decreasing or non-increasing) bounded activation function in the following corollary.

\begin{corollary}\label{corollary::negative-coreset}
Let $(P,w,\ball{\beta},f)$ be a general query spaces, of VC-dimension $O(d)$ such that $f(p,x)=\phi(p^Tx)$ for some  monotonic function   $\phi:\REAL\to \REAL$ and $P\subseteq \ball{\alpha}$.
%Suppose that $$s_{w_i}:P\to[0,\infty)$$ such that $s_w_i(p)\geq w_i(p)\sup_{x\in X}f(p,x)$.
Let $$ s(p) = \sup_{x\in X} |w(p)\phi(p^Tx)|$$ for every $p\in P$.
Let $c\geq1$ be a sufficiently large constant that can be determined from the proof, $t = \sum_{p \in P}s(p)$, and
\[
m\geq \frac{ct}{\varepsilon^2}\left(d\log t+\log\left(\frac{1}{\delta}\right)\right).
\]
Let $(C,u)$ be the output of a call to $\coreset(P,w,m,\phi,\beta)$; see Algorithm~\ref{coreset}. Then, $|C|\leq m$ and, with probability at least $1-\delta$,
\[
\forall x\in \ball{\beta}:\left|\sum_{p\in P}w(p)\phi(p^Tx)- \sum_{p\in C}u(p)\phi(p^Tx)\right| \leq \eps.
\]
\end{corollary}

\begin{proof}
We assume that $\phi$ is a non-decreasing function.
Otherwise, we apply the proof below for the non-decreasing function $\phi^* = -\phi$ and corresponding weight $w^*(p) = -w(p)$ for every $p\in P$. The correctness follows from $w(p)\phi(p^Tx)=w^*(p)\phi^*(p^Tx)$ for every $p\in P$.

Indeed, put $x \in \ball{\beta}$, and $\phi$ non-decreasing.
Hence,
{\small
\begin{align}
\nonumber&\left|\sum_{p\in P}w(p)\phi(p^Tx)- \sum_{p\in C}u(p)\phi(p^Tx)\right|\\
  \nonumber \leq &\left|\underset{\begin{subarray}{c}
                    p \in P\\
                  \nonumber   w(p)\phi(p^Tx) \geq 0
                  \end{subarray}}{\sum}w(p)\phi(p^Tx)-
                  \underset{\begin{subarray}{c}
                    p \in C\\
                   \nonumber u(p)\phi(p^Tx) \geq 0
                  \end{subarray}}{\sum}u(p)\phi(p^Tx)\right|\\
                \nonumber    +& \left| \underset{\begin{subarray}{c}
                    p \in P\\
                    w(p)\phi(p^Tx)< 0
                  \end{subarray}}{\sum}w(p)\phi(p^Tx) -
                  \underset{\begin{subarray}{c}
                    p \in C\\
           \nonumber          u(p)\phi(p^Tx) < 0
                  \end{subarray}}{\sum}u(p)\phi(p^Tx)\right|\\
       \nonumber \leq &\left| \underset{\begin{subarray}{c}
                    p \in P\\
          \nonumber           w(p)\phi(p^Tx)\geq 0
                  \end{subarray}}{\sum}w(p)\phi(p^Tx) - \underset{\begin{subarray}{c}
                    p \in C\\
          \nonumber           u(p)\phi(p^Tx)\geq 0
                  \end{subarray}}{\sum}u(p)\phi(p^Tx)\right|\\
         \nonumber          +& \left|\underset{\begin{subarray}{c}
                    p \in P\\
                   w(p)\phi(p^Tx)< 0
                  \end{subarray}}{\sum}|w(p)\phi(p^Tx)|
                  -\underset{\begin{subarray}{c}
                    p \in C\\
          \nonumber           u(p)\phi(p^Tx)< 0
                  \end{subarray}}{\sum}|u(p)\phi(p^Tx)|\right|
\end{align}
}
The first inequality in the the proof is obtained by separating each sum into points with positive and negative weights and applying Cauchy-Schwarz inequality. The second inequality is obtained by bounding points with positive and negative weights separately using  Theorem~\ref{lem:sensitivity}.
\end{proof}

\begin{figure*}%[hb!]
\centering
\includegraphics[width=0.8\textwidth]{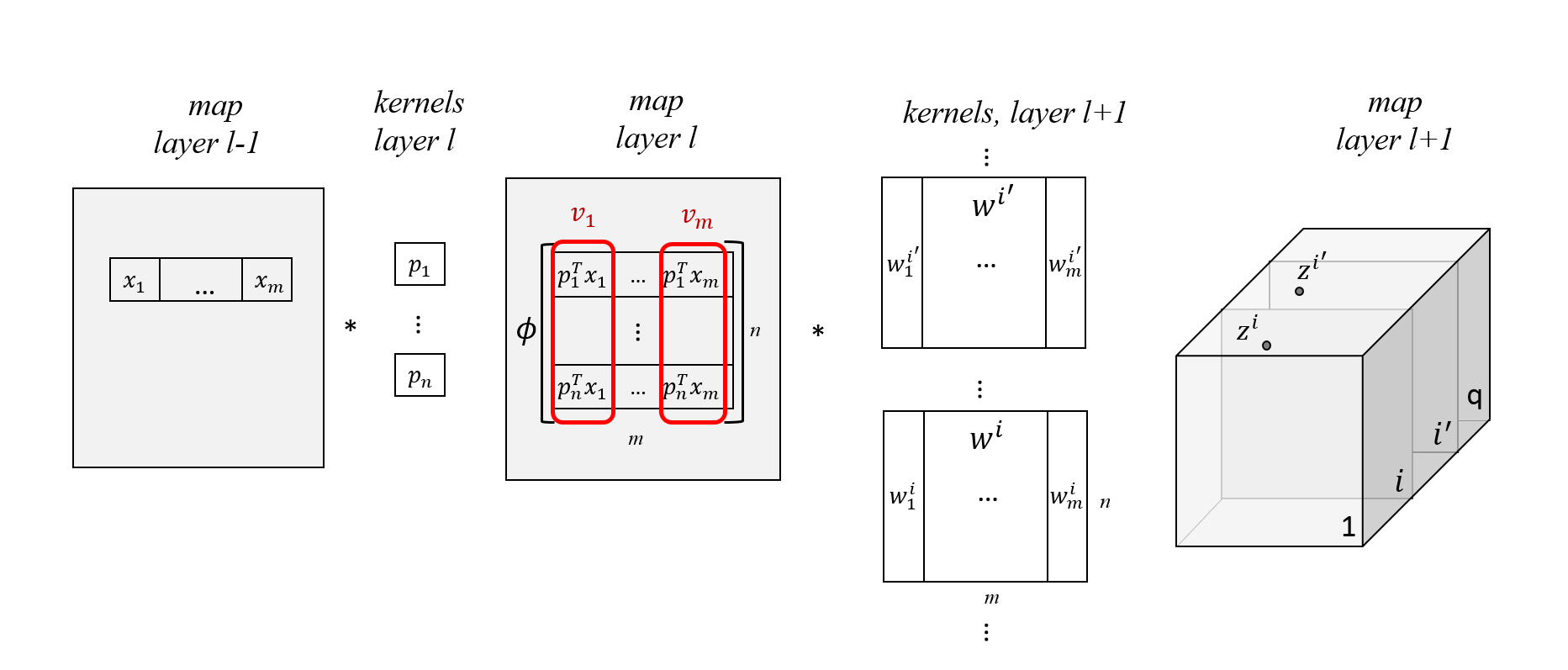}
\caption{Illustration for the construction of channel pruning coreset. The filters in layer $l$ are one-dimensional for simplicity. The same construction applies for any tensor.}\label{fig:chan_pruning_illust}
\end{figure*}

\subsubsection{From Coreset per Neuron to Coreset per Layer}\label{sec:dense_layer_coreset}
%In this section we apply Algorithm~1 on each neuron in a specific layer $i+1$. This is by computing a coreset of its inner nodes from layer $i$. However, it might be that each layer in level $i+1$ pick a different subset (coreset) in layer $i$.
Applying Algorithm~\ref{coreset} to each neuron in a layer $i+1$ could result in the situation that a neuron in layer $i$ is selected to the coreset of some neurons in layer $i+1$, but not to others. In this situation, it cannot be removed. To perform neuron pruning, every neuron in layer $i + 1$ should select the same neurons for its coreset, maybe with different weights. Thus, we wish to compute a single coreset for multiple weighted sets that are different only by their weight function. Each such a set represents a neuron in level $i+1$, which includes $k$ neurons. Algorithm~2 and Corollary~\ref{corollary::double-coreset}  show how to compute a single coreset for multiple weighted sets. Figure~\ref{fig:schem_single_n} provides an illustration of the layer pruning on a toy example.

\begin{algorithm}[!htb]
\caption{\small $\lcoreset(P,w_1,\cdots,w_k,m,\phi,\beta)$  }\label{coreset2}
{\begin{tabbing}
\textbf{Input:} \quad\quad \= Weighted sets $(P,w_1),\cdots,(P,w_k)$, \\\>an integer (sample size) $m \geq 1$,
\\ \>an (activation) function $\phi:\REAL\to[0,\infty)$,
\\ \> an upper bound $\beta>0$.\\
\textbf{Output:} \>A weighted set $(C,u)$; see Theorem~\ref{lem:sensitivity}.
\end{tabbing}}
%\vspace{-0.3cm}
%\For{every $p\in P$}
%{
%$\displaystyle s(p):= $
%}
\For{every $p\in P$}
{
$\pr(p):=\displaystyle\frac{\max_{i\in[k]} w_i(p)\phi(\beta \norm{p})}{\sum_{q\in P}\max_{i\in[k]}w_i(q)\phi(\beta \norm{q})}$\\
$u(p):=0$
}
$C \gets \emptyset$\\
\For{$m$ iterations}
{
Sample a point $q$ from $P$ such that $p\in P$ is chosen with probability $\pr(p)$.\\
$C:=C\cup\{q\}$\\
$\forall i\in[k]: u_i(q):= u_i(q)+\displaystyle\frac{w_i(q)}{m\cdot\pr(q)}$
}
\Return $(C,u_1,\cdots,u_k)$
\end{algorithm}

%We use the following observation of how to compute a single coreset for multiple weighted sets that are different only by their weight function. Each such a set represents a neuron in level $i+1$ of $k$ neurons. The proof follows directly from Theorem~\ref{cor11}.
\begin{corollary}[Coreset per Layer\label{corollary::double-coreset}]
Let $(P,w_1,\ball{\beta},f), \dots ,(P,w_k,\ball{\beta},f)$ be $k$ query spaces, each of VC-dimension $O(d)$ such that $f(p,x)=\phi(p^Tx)$ for some non-decreasing $\phi:\REAL\to[0,\infty)$ and $P\subseteq \ball{\alpha}$.
%Suppose that $$s_{w_i}:P\to[0,\infty)$$ such that $s_w_i(p)\geq w_i(p)\sup_{x\in X}f(p,x)$.
Let $$ s(p) = \max_{i\in[k]}\sup_{x\in X} w_i(p)\phi(p^Tx)$$ for every $p\in P$.
Let $c\geq1$ be a sufficiently large constant that can be determined from the proof, $t = \sum_{p \in P}s(p)$
\[
m\geq \frac{ct}{\varepsilon^2}\left(d\log t+\log\left(\frac{1}{\delta}\right)\right).
\]
Let $(C,u_1,\cdots,u_k)$ be the output of a call to $\coreset(P,w_1,\cdots,w_k,m,\phi,\beta)$; see Algorithm~\ref{coreset2}. Then, $|C|\leq m$ and, with probability at least $1-\delta$,
\begin{align}
\nonumber&\forall i\in[k],x\in \ball{\beta}:\\
\nonumber&\left|\sum_{p\in P}w_i(p)\phi(p^Tx)- \sum_{p\in C}u_i(p)\phi(p^Tx)\right| \leq \eps.
\end{align}
\end{corollary}
The proof follows directly from the observation in Theorem~\ref{cor11} that $s(p) \geq w(p)\sup_{x\in X}f(p,x)$.

%\textbf{Support for negative weights }is possible by simply applying Algorithm~2 independently for the positive and negative edges of each neuron; see e.g.~\citet{baykal}.

 %add positive and negative weigths
 %add figure
 %add coreset for several neurons.

\subsection{From Neural Pruning to Channel Prunning in CNNs}\label{sec:conv_layer}
%Let $x_1, \dots, x_m \in \real^d$, let $p_1, \dots, p_n \in \real^d$ be a set of kernels in convolutional layer $l$ and let  $w_1, \dots, w_k \in \real^{m \times n}$ be a set of kernels in convolutional layer $l + 1$. Denote $v_i := (x_i^T p_1, \dots, x_i^T p_n)$.
%Based on the results in Section~\ref{sec:analysis}, we know how to approximate $\sum_{i \in [n]}w_i \phi(p_i ^T x)$ using coreset. Using this coreset we will approximate the output of the convolutional layer $l + 1$. The $i$-th linear output of the layer $l + 1$ is $w_i ^T \phi(v)$ such that $v$ is a concatenation of $v_1, \dots, v_m$, and $\phi(v) = (\phi(v[1]), \dots, \phi(v[m \times n]))$.
%$w_i ^T \phi(v) = \sum_{j}w_{ij}^T \phi(v_j)$. Using Lemma \ref{corollary::double-coreset} for computing coreset for multiple set, we can approximate $w_i ^T \phi(v)$. By applying this lemma again for every linear output in layer $l + 1$, we get dimensional pruning (channel pruning) of layer $l$ and $l + 1$.
Let $x_1, \dots, x_m \in \real^d$ denote $m$ adjacent segments in the map of layer $l-1$ (see illustration in Figure~\ref{fig:chan_pruning_illust}).
Let $p_1, \dots, p_n \in \real^d$ be a set of kernels in convolutional layer $l$ and let  $w^1, \dots, w^q \in \real^{m \times n}$ be a set of kernels in convolutional layer $l + 1$. Let $v_j:=[p_1^T x_j,\dots, p_n^T x_j]$ be a linear part of the map in layer $l$ that corresponds to the input segment $x_j$ (see Figure~\ref{fig:chan_pruning_illust} for illustration) and let $v$ denote a concatenation of $v_1,\dots,v_m$. Consequently, $\phi(v)$ is $m\times n$ patch in the map of layer $l$, which is obtained by applying $n$ kernels of layer $l$ to input $x_1, \dots, x_m$.

We start by approximating the linear part of element $z^i$ in channel $i$ of the map in layer $l+1$ using a coreset of kernels in layer $l$. Under the above notations,
\begin{align}\label{eq_chan}
 z^i&=\langle w^i,\phi(v)\rangle=\sum_{j=1}^m(w^i_j)^T\phi(v_j)\\
 \nonumber &=\sum_{j=1}^m \left( \sum_{k=1}^n w^i_{jk}\phi(p_k^T x_j)\right),
\end{align}
where $w^i$ is the corresponding kernel in layer $l+1$.  Theorem~\ref{lem:sensitivity} and Corollary~\ref{corollary::negative-coreset} show how to approximate each of $\sum_{k=1}^n w^i_{jk}\phi(p_k^T x_j)$, but these approximations should use the same subset of kernels $\{p_k\}$. To this end, we define $m$ weighted sets $(P,w^i_1),\dots,(P,w^i_m)$, where $P=\{p_1,\dots,p_n\}$ and compute a single coreset for them using Algorithm 2. We note that here each weighted set operates on a different query $x_j$. Since coreset holds for any query, we can still use Algorithm 2 to find a single coreset of $P$ to approximate $z^i$.

Next, we consider another element $\tilde{z}^i$ in the same channel $i$ of the linear part of the map in layer $l+1$. Similarly to Eq.~\ref{eq_chan}, $$\tilde{z}^i=\sum_{j=1}^m \left( \sum_{k=1}^n w^i_{jk}\phi(p_k^T \tilde{x}_j)\right),$$
where $\tilde{x}_j$, $j=1, \dots,m$ is a different part of the map in layer $l-1$. $\tilde{z}^i$ differs from $z^i$ only in the query, thus we can define the same weighted sets for $z^i$ and $\tilde{z}^i$. Same argument applies to all elements in the channel, thus same coreset will hold for approximating every element in channel $i$ of the map of layer $l+1$.

Finally, we consider an element in a different channel $i'$ of the map in layer $l+1$. Its linear part is defined as follows,
$$z^{i'}=\sum_{k=1}^n w^{i'}_{jk}\phi(p_k^T x_j)$$ and its approximation must choose the same subset of kernels $\{p_k\}$ as  the approximation of elements in channel $i$. This case is similar to approximating multiple neurons proposed in Section~\ref{sec:dense_layer_coreset}. Following the derivation for dense layer pruning, we define weighted sets for each channel $i=1,\dots,q$ as $\{(P,w^i_1),\dots,(P,w^i_m)\}$ and apply Algorithm 2 to the union of the weighted sets associated with all channels in the map: $\bigcup_{i=1}^q \{(P,w^i_1),\dots,(P,w^i_m)\}$.

\begin{figure*}
\centering
\subfloat[\label{fig:mnist-size}]{
  \includegraphics[width=0.32\textwidth]{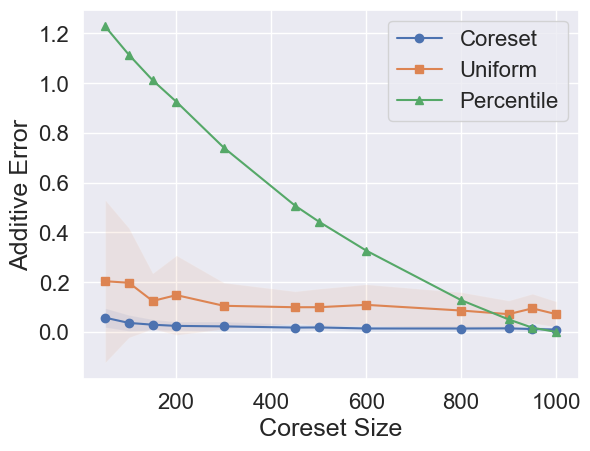}}
\subfloat[\label{fig:mnist-sparsity}]{
    \includegraphics[width=0.32\textwidth]{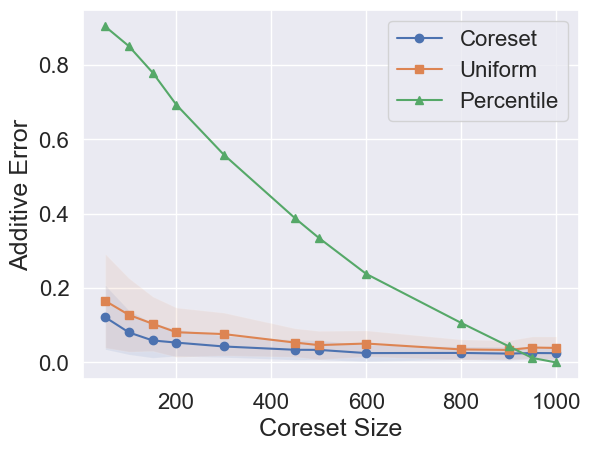}}
\subfloat[\label{fig:mnist-real}]{
    \includegraphics[width=0.32\textwidth]{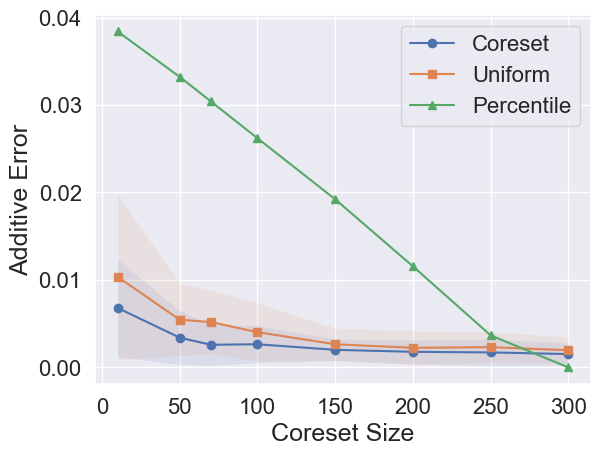}}
  \caption{Approximation error of a single neuron on MNIST dataset across different coreset sizes. The weights of the points in (a) are drawn from the Gaussian distribution, in (b) from the Uniform distribution and in (c) we used the trained weights from LeNet-300-100. Our coreset, computed by Algorithm~1 and Corollary~8, outperforms other reduction methods.}
  \label{fig:plot1}
\end{figure*}

\section{Experiments}
Our first set of experiments focuses on compression of dense layers. We start with the analysis of the proposed \emph{Coreset for Neuron} algorithm and then compare the performance of the \emph{Coreset per Layer} algorithm with other coreset-based methods. We then continue with testing our coreset-based framework for dense layers on  benchmark architectures and data sets. We conclude this part of the experiments with an ablation analysis that evaluates the contribution of different parts of our framework.

The second set of experiments focuses on channel pruning. We test our coreset-based channel pruning on the benchmark architectures and data sets and compare it to previous methods that use heuristics for model approximation. Finally, we compare the speed of the fine-tuning that updates the model approximated using our coreset-based channel pruning with training the same architecture from scratch. This experiment demonstrates the affect of accurate approximation on the construction time. Specifically, it shows very significant savings in fine-tuning a well-approximated compressed model compared to training a compressed model from scratch as was suggested in~\cite{rethinking}.

%The experiments were implemented in PyTorch~\citep{PyTorch} on a Linux Machine using an Intel Xeon, 32-core CPU with 3.2 GHz, 256 GB of RAM and Nvidia TitanX and Quadro M4000 GPUs.

\subsection{Coreset per Neuron}
\subsubsection{Approximation error vs. Coreset Size} We analyzed the empirical trade-off between the approximation error of ReLU neuron and the size of its coreset, constructed by Algorithm~1 and Corollary~\ref{corollary::negative-coreset}. We compared our coreset  to \emph{uniform sampling}, which also implements Algorithm~1, but sets the probability of a point to  $1/n$ ($n$ is the size of the full set), and to \emph{percentile}, which deterministically retains the inputs with the highest norms (note that in percentile the points are not weighted). We ran three tests, varying the distribution of weights. In the first and second tests (Figure~\ref{fig:plot1}, (a) and (b)) the weights were drawn from the Gaussian and Uniform distributions respectively. The total number of neurons was set to 1000. We selected subsets of neurons of increasing sizes from 50 to 1000  with a step of 50. In the third test (Figure~\ref{fig:plot1}, (c)) we used the trained weights from the first layer of Lenet-300-100 including 300 neurons (LeNet-300-100 network comprises two fully connected hidden layers with 300 and 100 neurons correspondingly and ReLu activations, trained on MNIST~\cite{MNIST} train set). We varied the coreset size from 50 to 300 with a step 50. To evaluate the approximation error, we used images from MNIST test set as queries. Each point in the plot was computed by  1) running the full network and the compressed network (with corresponding compression level) on each image $x$ in the test set, 2) computing additive approximation error $\left|\sum_{p\in P}w(p)\phi(p^Tx)- \sum_{p\in C}u(p)\phi(p^Tx)\right|$,  3) averaging the resulting error over the test set. In all three tests, our coresets outperformed the tested methods across all coreset sizes.
\begin{figure*}
\centering
\subfloat[\label{fig:plot_iclr}]{\includegraphics[width=0.37\textwidth]{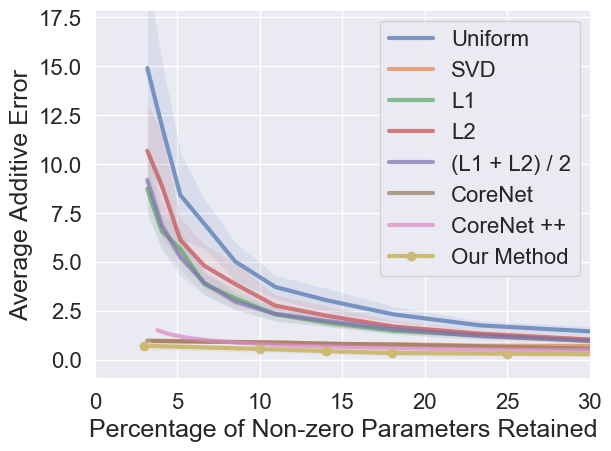}}
\subfloat[\label{fig:plot_iclr2}]{\includegraphics[width=0.37\textwidth]{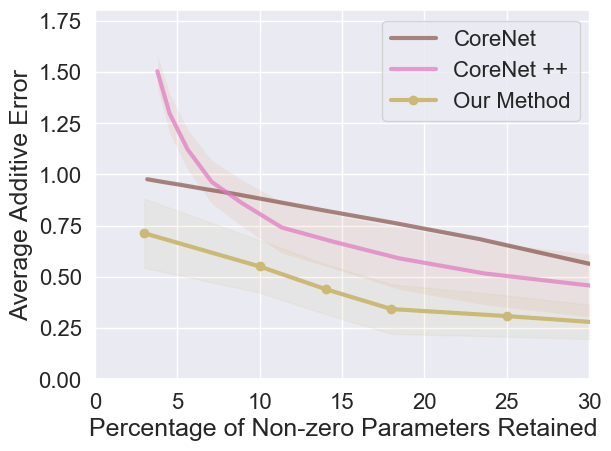}}
  \caption{Average accuracy of various algorithms on LeNet-200-105 on MNIST dataset across different sparsity rates. Plot (a) shows the results of all tested methods. Plot (b) focuses on the three top methods. Our method, which constructs neural coresets by applying the Algorithm~2 and Corollary~8 in a layer-by-layer fashion, outperforms other coreset-based algorithms.}
  \label{fig:plot}
\end{figure*}

\subsubsection{Comparison with Other Coreset-Based Methods}
We compared the average approximation error vs. compression rates of our neural pruning coreset with several other well-known algorithms listed blow.
\begin{itemize}
   \item\textbf{Baselines:} uniform sampling, percentile (which deterministically retains the inputs with the highest norms), and Singular Value Decomposition (SVD);
   \item\textbf{Schemes for matrix sparsification:} based on L1 and L2 norms and their combination~\citep{drineas2011note,achlioptas2013matrix,KunduD14};
   \item \textbf{Sensitivity sampling:} CoreNet and CoreNet++~\citep{baykal}.
 \end{itemize}
The tests were performed on LeNet-200-105 architecture, which includes two fully connected hidden layers with 200 and 105 neurons correspondingly and ReLu activations, trained on MNIST~\cite{MNIST}.  We computed the average error of the tested algorithms over the samples of the MNIST test set after performing each test ten times. We measured the corresponding average approximation error as defined in~\citep{baykal}:
$$error_{\mathcal{P}_{test}}=\frac{1}{\mathcal{P}_{test}}\sum_{x\in \mathcal{P}_test}\|\phi_{\hat{\theta}}(x)-\phi_{\theta}(x)\|_1,$$
where $\phi_{\hat{\theta}}(x)$ and $\phi_{\theta}(x)$ are the outputs of the approximated and the original networks respectively.
The results are summarized in Figure~\ref{fig:plot}. As expected, all algorithms perform better with lower compression, but our algorithm outperforms other methods, especially for high compression rates.

\begin{table}
\centering
\begin{tabular}{|l|cc|c|}
\hline
Network & Error(\%) & \# Parameters & Compression \\
& & &Ratio \\\hline \hline
LeNet-300-100  & 2.16 & 267K &  \\
LeNet-300-100 & \textbf{2.03} & \textbf{26K} & \textbf{90\%}\\
Pruned & & &\\\hline
VGG-16& 8.95 & 1.4M &  \\
VGG-16 Pruned & \textbf{8.16} & \textbf{350K} & \textbf{75\%}
\\ \hline
\end{tabular}
\caption{Empirical evaluations of our coresets on existing architectures for MNIST and CIFAR-10. Note the improvement of accuracy in both cases.}
\label{table:experiments}
\end{table}

\subsection{Neural Pruning via Coreset per Layer}
We tested the proposed framework for neural pruning via coresets on dense layers of two popular models: LeNet-300-100 (fully-connected network described above) on MNIST~\citep{MNIST}, and VGG-16~\citep{Simonyan14c} on CIFAR-10~\citep{krizhevsky2009learning}. In addition to the convolutional and pooling layers, VGG-16~\citep{Simonyan14c} includes 3 dense layers -- the first two with 4096 neurons and the last with 1000 neurons (we applied our algorithm for neural pruning to the dense layers). In both networks, we first applied neural pruning using Coreset per Layer (Algorithm~2), and then fine-tuned the remaining weights until convergence.

Our method was able to prune roughly $90\%$ of the parameters of LeNet-300-100 network without any accuracy loss -- in fact, it slightly improved the classification accuracy.  After compressing the dense layers of VGG-16 network by roughly $75\%$, its accuracy also showed slight improvement. We summarize our findings in Table~\ref{table:experiments}.

\subsubsection{Ablation Analysis}
\begin{figure*}
\centering
\subfloat[]{\includegraphics[height=1.6in]{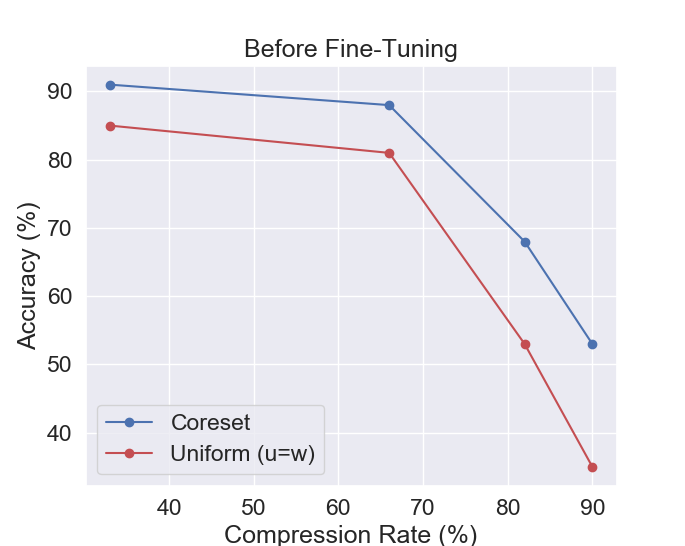}}
\subfloat[]{\includegraphics[height=1.6in]{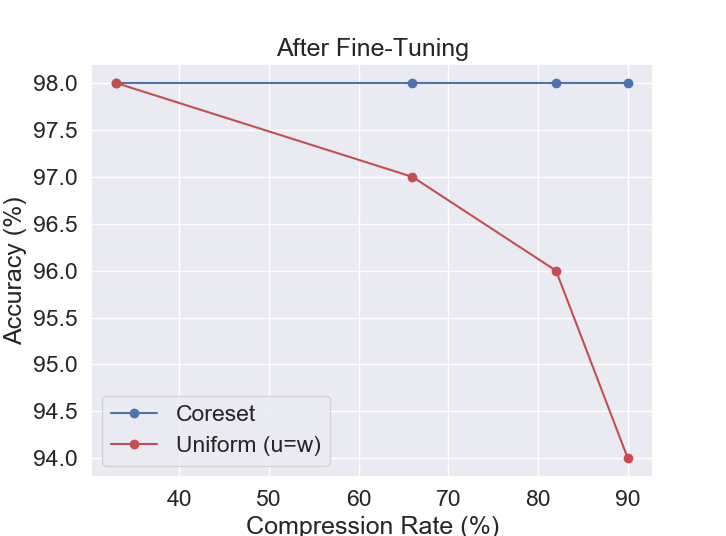}}
  \caption{Average accuracy over 5 runs of the proposed framework and the uniform baseline on LeNet-300-100 on MNIST dataset across different compression rates. Plot (a) shows the results before fine-tuning, plot (b)-after fine-tuning. The fine-tuning was done until convergence. Fine-tuning of the uniform sampling almost doubles in time compared to the fine-tuning of the coreset.}
  \label{fig:ablation}
\end{figure*}

The proposed compression framework includes for every layer, a selection of neurons using Algorithm~2, followed by fine-tuning. We performed the following ablation analysis to evaluate the contribution of different parts of our framework on  LeNet-300-100  trained on MNIST.
First, we removed the fine-tuning, to test the improvement due to Algorithm~2 over the uniform sampling. Figure~\ref{fig:ablation}, (a) shows the classification accuracy without fine-tuning as a function of the compression rate. Figure~\ref{fig:ablation}, (b) shows that fine-tuning improves both methods, but the advantage of the coreset is still apparent across almost all compression rates and it increases at the higher compression rates. Note that the model selected by the coreset can be fine-tuned to 98\% classification accuracy for any compression rate, while the model chosen uniformly cannot maintain the same accuracy for high compression rates.

These results demonstrate that our coreset algorithm provides better selection of neurons compared to uniform sampling. Moreover, it requires significantly less fine-tuning: fine-tuning until convergence of the uniform sampling took close to 2 epochs, while fine-tuning of our method required about half of that time.

\subsection{Channel Pruning}
\label{sec_experiments_CP}
We ran our experiments on three benchmark models: VGG~\citep{Simonyan14c} and ResNet56 \citep{resnet} on CIFAR-10~\citep{krizhevsky2009learning}, and ResNet50~\citep{resnet} on ILSVRC2012~\citep{ILV}. In all our experiments, we applied the proposed coreset framework to approximate the original network using the small target architecture. Specifically, we  compressed in one shot the convolutional layers of the original network using the coreset for channel pruning detailed in Section~\ref{sec:conv_layer} to the size, defined by the target small architecture. The results below show that channel pruning via coresets is comparable with the prior work in terms of the size and accuracy of the final model, but the compressed network provides better approximation of the original network, consequently it requires less fine-tuning, and this approximation holds for any input.

%\begin{enumerate}
%    \item Network Slimming ~\citep{slim}
%    \item CCP ~\citep{collab}
%    \item Channel Pruning \citep{ChannelPF}
%    \item Soft Pruning \citep{soft}
%    \item AMC \citep{automl}
%    \item Norm-Based \citep{nec}
%    \item ThiNet \citep{thin}
%    \item FPGM \citep{median-pruning}
%\end{enumerate}

%In all experiments we used ReLU networks and for every layer we applied channel pruning. The fine-tuning setting for VGG-19 is the same as \citep{network_sliming} and the setting for ResNet-56 is the same as \citep{collab}. The experiments were implemented in PyTorch~\citep{PyTorch} on a Linux Machine using an Intel Xeon, 8-core CPU with 2.1 GHz, 0.5 TB of RAM and two Nvidia Tesla-100 GPUs\ignore{\footnote{We will make our code available upon acceptance or on reviewers' request.}}

\begin{table}
\centering
\begin{tabular}{|c|c||c|c|}
\hline
Layer&Width&Layer&Width\\
\hline
1&49&9&114\\
2& 64&10&41\\
3&128&11&24\\
4& 128&12&11\\
5& 256&13&14\\
6& 254 & 14&13\\
7& 234&15&19\\
8& 198&16&104\\
\hline
\end{tabular}
\caption{VGG-19 compressed architecture}\label{VGG_comr_arch}
\end{table}

\begin{table*}
\centering
\begin{tabular}{|l|ccc|}
\hline
Pruning Method & Baseline(\%) & Small Model Acc.(\%)& Compression Ratio \\\hline \hline
Unstructured Pruning~\cite{Han}& 6.5 & 6.48 & 80\%\\
Structured Pruning~\cite{slim} & 6.33 & 6.20 & 70\% \\
Pruning via Coresets &6.33 &6.02 & 70\% \\
\hline
\end{tabular}
\caption{Pruning of VGG-19 (20M parameters) for CIFAR-10}
\label{table:comparison-vgg}
\end{table*}

\begin{table*}%[!htb]
\centering
\begin{tabular}{|l|ccc|}
\hline
Pruning Method & Baseline(\%) & Small Model Acc.(\%)& Compression Ratio \\\hline \hline
Channel Pruning~\cite{ChannelPF} &7.2& 8.2 &  39\%
\\
%AMC~\cite{automl} &7.2& 7.11 &  39\%
AMC~\cite{automl} &7.2& 8.1 &  40\%
\\
%Soft Pruning~\cite{soft} & 6.48 &  40\%
Soft Pruning~\cite{soft} & 6.41 & 6.65 & 40\%
\\
%CCP~\citep{collab}& 6.29 &  40\%
CCP~\citep{collab} &6.5& 6.42 &  40\%
\\
Pruning via Coresets & 6.21 &7.0 &  40\%
\\\hline
\end{tabular}
\caption{Channel Pruning of ResNet-56 (with 860K parameters) on CIFAR-10}
\label{table:comparison-resnet}
\end{table*}

\begin{table*}%[!htb]
\centering
\begin{tabular}{|l|ccc|}
\hline
Pruning Method & Baseline Top-1(\%) & Small Model Top-1 Acc.(\%)& Compression Ratio \\\hline \hline
Soft Pruning~\cite{soft} & 23.85 & 25.39 & 30\%
\\
CCP~\citep{collab} & 23.85& 24.5 &  ~35\%
\\
FPGM~\citep{median-pruning}& 23.85& 25.17 & 40\%
\\
ThiNet-70~\citep{thin} &27.72& 26.97 &  30\%
\\
ThiNet-50~\citep{thin} &27.72&28.0 &  50\%
\\
Pruning via Coresets&23.87 & 25.11 &  40\%\\
\\\hline
\end{tabular}
\caption{Channel Pruning of ResNet-50 (with 26M parameters) on ILSVRC-2012}
\label{table:comparison-resnet50}
\end{table*}

\subsubsection{Compressing VGG-19}
We used pytorch implementation\footnote{https://github.com/Eric-mingjie/network-slimming} of VGGNet-19 network for CIFAR10 from~\cite{slim}  with  about 20M parameters as our baseline model. The target architecture of the small network\footnote{https://github.com/foolwood/pytorch-slimming} that corresponds to 70\% compression ratio and to the reduction of the parameters by roughly $88\%$ is shown in Table~\ref{VGG_comr_arch}. The results are summarized in Table~\ref{table:comparison-vgg}.

The approximation of the baseline VGG-19 using coresets before fine-tuning provided 12\% improvement of accuracy compared to~\cite{slim}. Consequently, our approximation needs less fine-tuning time to reach the original (or even improved) accuracy. Since our framework is one-shot, its coreset construction is very fast and it requires less fine-tuning (due to more accurate approximation), our framework is significantly more efficient that previous methods.

\subsubsection{Compressing ResNets}
We tested our method on two models: ResNet-56 for CIFAR-10 and ResNet-50 on ILSVRC-2012. The baseline ResNet-56 \citep{resnet} includes 56 convolutional layers with batch-normalization and dense layer with 64 neurons, in total about 860K parameters. The baseline ResNet-50~\cite{resnet} includes 50 convolutional layers with batch-normalization and dense layer with 2048 neurons, in total about 26M parameters.
We applied the same 40\% compression to every convolutional layer using the coreset algorithms for channel pruning in both networks. We then fine-tuned the compressed network.

We decreased the number of parameters in ResNet-56 by roughly 55\% and in ResNet-50 by by roughly 62\%. We compare our results for ResNet-56 with previous methods in Table~\ref{table:comparison-resnet} and for ResNet-50 in Table~\ref{table:comparison-resnet50}.
Our algorithm shows comparable accuracy and compression ratio with previous methods, but it has the advantage of simple and efficient construction and theoretical guarantees for any input.

\begin{figure*}%[!htb]
\centering
\subfloat[]{\includegraphics[height=1.7in]{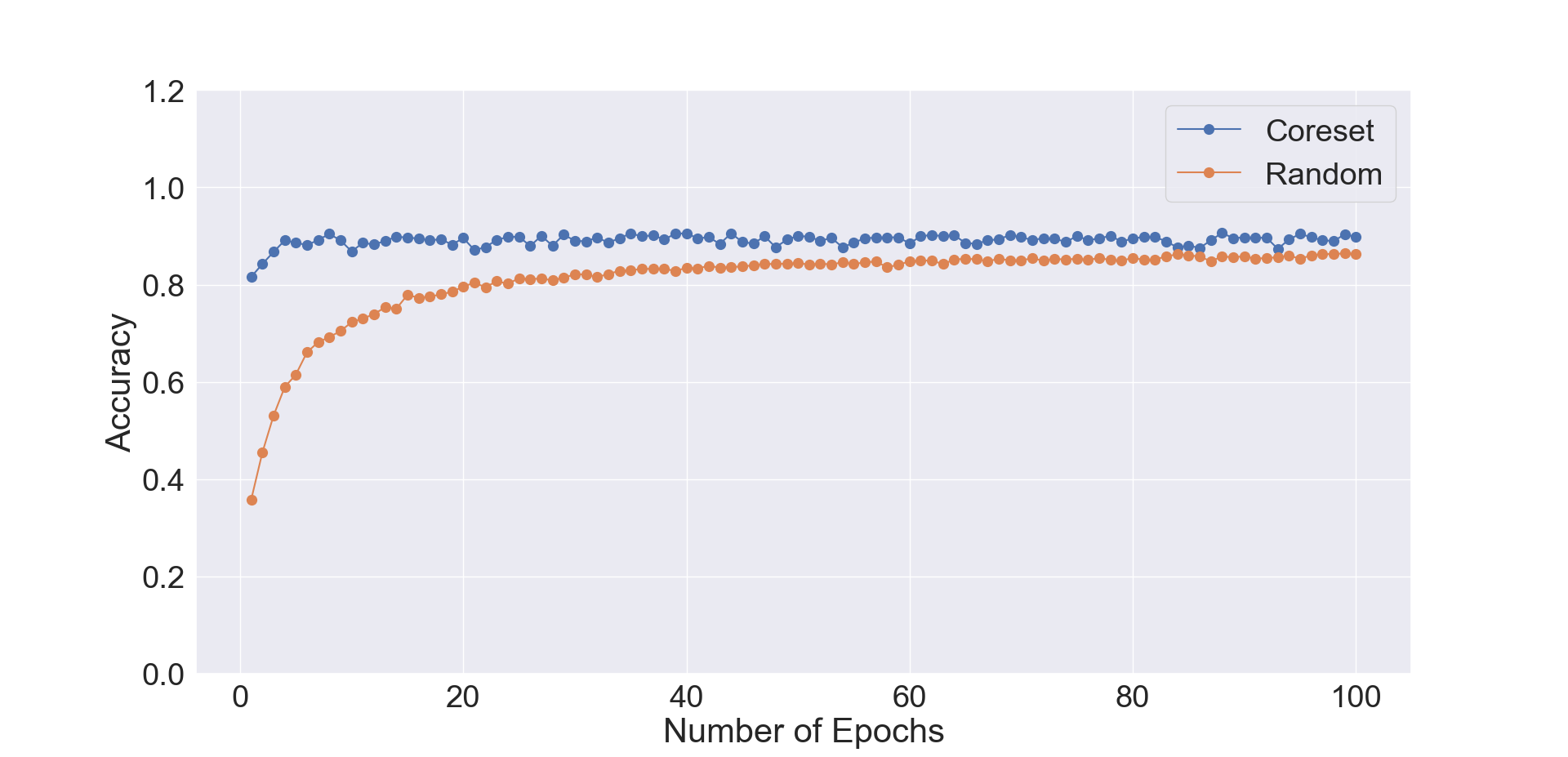}}
\subfloat[]{\includegraphics[height=1.7in]{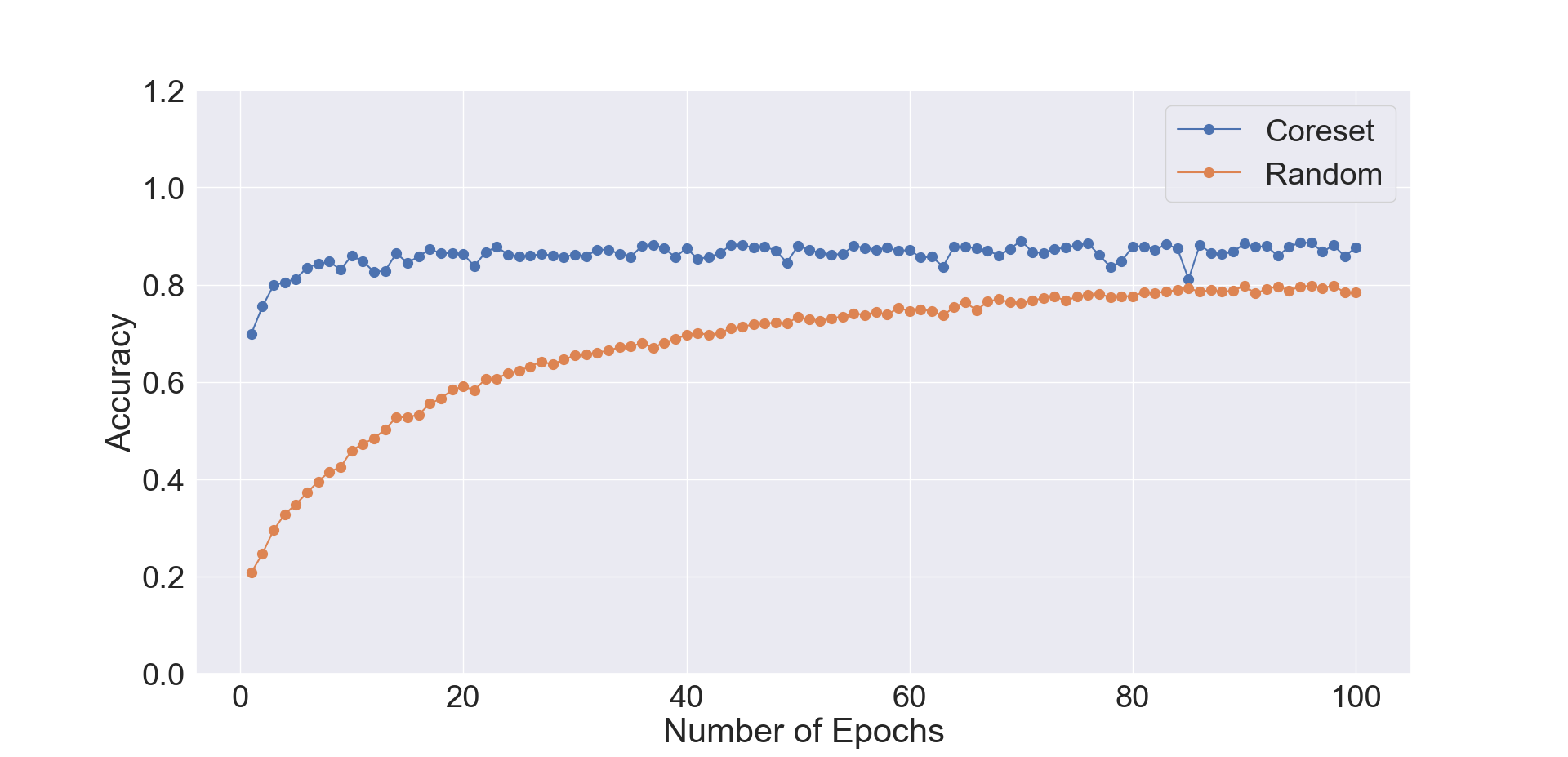}}\\
\subfloat[]{\includegraphics[height=1.7in]{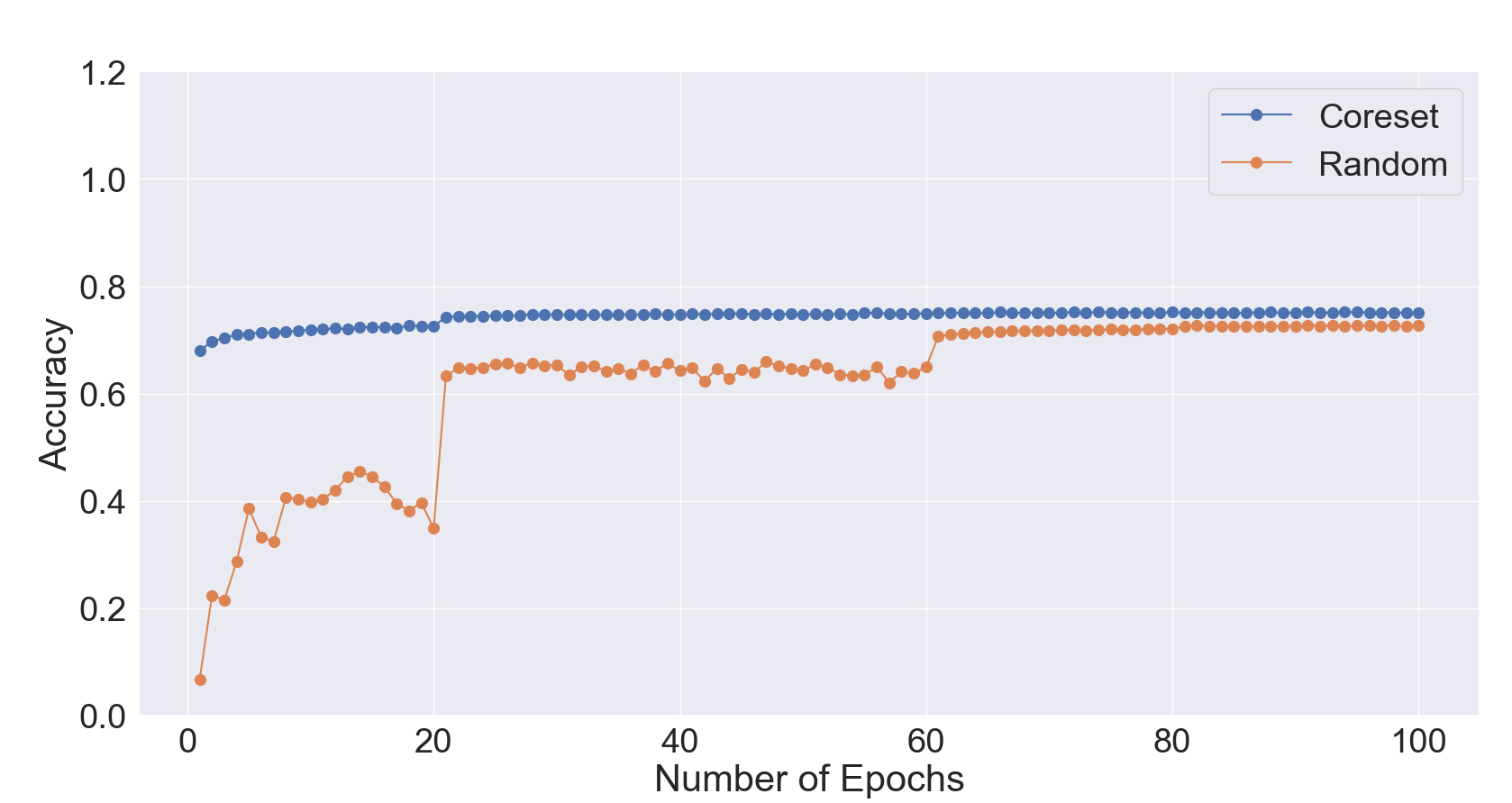}}
  \caption{Accuracy of the proposed framework and the random initialization during fine-tuning. Plot (a) shows the results for VGG-19 with CIFAR-10, plot (b) shows the results for ResNet-56 with CIFAR-10 and plot (c) shows the results for ResNet-50 with ImageNet.}
  \label{fig:tuning}
\end{figure*}
\subsection{The value of approximation in channel pruning}
Recently, \citep{rethinking, rewind} challenged some common beliefs about fine-tuning of compressed models and how it should be done. The claim in \citep{rethinking} was that when using a predefined architecture in structured compression, training the small architecture from random initialization is the best solution. Thus the only question one need to ask in network compression is what is the compressed architecture. Even though fully trained network could match the accuracy of a fine-tuned network, there is an advantage of the approximated network in terms of construction time: fine-tuning a well-approximated network is significantly faster than training the same architecture from scratch. The plots showing network accuracy over fine-tuning (Figure \ref{fig:tuning}) support our claim for all tested architectures.  %The gap in the performance is more apparent for the larger data set, proving the value of  approximation in network pruning.

\section{Conclusion}
We proposed the first structured pruning algorithm with provable trade-off between the compression rate to the approximation error for any future test sample. We base our compression algorithm on coreset framework and construct corsets for most common activation functions. Our tests on ReLU  dense networks and popular CNN architectures show high compression rates with comparable accuracy, and our theory guarantees the worst case accuracy vs. compression trade-off for any future test sample, even an adversarial one. The construction time of the proposed framework including fine-tuning is very fast, rendering our method very attractive to practitioners.  In future work, we plan to derive coreset of a full network and extend it to other architectures, such as transformer.

%\appendices

%\section{}

%\section*{Acknowledgment}




%\bibliographystyle{IEEEtran}
\bibliographystyle{abbrv}
\bibliography{nueralCoreset}


%\end{IEEEbiography}


%\newpage



%\vfill


\end{document}